\documentclass{article}

     \usepackage[preprint,nonatbib]{neurips_2020}

\usepackage[utf8]{inputenc} 
\usepackage[T1]{fontenc}    
\usepackage{hyperref}       
\usepackage{url}            
\usepackage{booktabs}       
\usepackage{amsfonts}       
\usepackage{nicefrac}       
\usepackage{microtype}      
\usepackage{xspace}
\usepackage{graphicx}
\graphicspath{{./NeurIPS_style}}

\newcommand{\Pasithea}{{P}{\footnotesize ASITHEA}\xspace}
\newcommand{\Pasitheaa}{{P}{\footnotesize ASITHEA}\xspace}
\newcommand{\describe}{{\footnote{\Pasithea is the goddess of relaxation, meditation, hallucinations, and wife of Hypnos, the god of sleep.}}{\scriptsize}\xspace}

\newcommand*{\selfies}{{S}{\footnotesize ELFIES}\xspace}
\title{Deep Molecular Dreaming:\\Inverse machine learning for de-novo molecular design and interpretability with surjective representations}

\author{
Cynthia Shen $^{1,\star}$ \\
cynt.shen@mail.utoronto.ca \\
\And 
Mario Krenn$^{1,2,3,\star}$ \\
mario.krenn@utoronto.ca \\
\And
Sagi Eppel$^{1,2,3}$ \\
sagieppel@gmail.com \\
\And
Al\'{a}n Aspuru-Guzik$^{1,2,3,4}$\\
alan@aspuru.com \\
\And
\\
  $^1$Department of Computer Science, University of Toronto, Canada.\\
  $^2$Chemical Physics Theory Group, Department of Chemistry, University of Toronto, Canada.\\
  $^3$Vector Institute for Artificial Intelligence, Toronto, Canada. \\
  $^4$Canadian Institute for Advanced Research (CIFAR) Lebovic Fellow, Toronto, Canada. \\
  $^{\star}$These authors contributed equally
}

\begin{document}

\maketitle

\begin{abstract}
Computer-based de-novo design of functional molecules is one of the most prominent challenges in cheminformatics today. As a result, generative and evolutionary inverse designs from the field of artificial intelligence have emerged at a rapid pace, with aims to optimize molecules for a particular chemical property. These models 'indirectly' explore the chemical space; by learning latent spaces, policies, distributions or by applying mutations on populations of molecules. However, the recent development of the \selfies \cite{krenn2020self} string representation of molecules, a surjective alternative to SMILES, have made possible other potential techniques. Based on \selfies, we therefore propose \Pasitheaa, a direct gradient-based molecule optimization that applies inceptionism \cite{mordvintsev2015inceptionism} techniques from computer vision. \Pasithea exploits the use of gradients by directly reversing the learning process of a neural network, which is trained to predict real-valued chemical properties. Effectively, this forms an inverse regression model, which is capable of generating molecular variants optimized for a certain property. Although our results are preliminary, we observe a shift in distribution of a chosen property during inverse-training, a clear indication of \Pasithea's viability. A striking property of inceptionism is that we can directly probe the model's \textit{understanding} of the chemical space it was trained on. We expect that extending \Pasithea to larger datasets, molecules and more complex properties will lead to advances in the design of new functional molecules as well as the interpretation and explanation of machine learning models. 
\end{abstract}

\section{Introduction}
\begin{figure}[t]
\centering
\includegraphics[width=\textwidth]{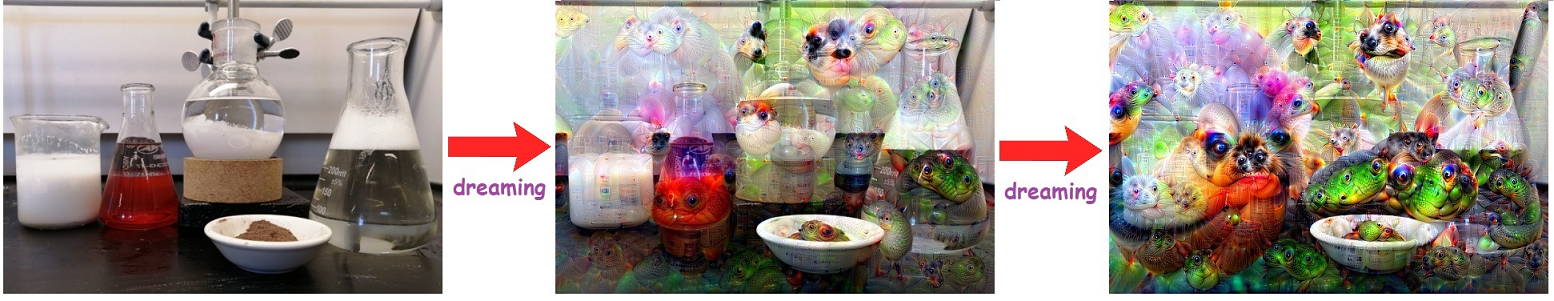}
\caption{Deep dreaming is well-known for creating new dream-like images \cite{simonyan2014very}. To generate this image, we used the following github repo \cite{linden2019github}.}
\label{fig:glassware}
\end{figure}

The de-novo design of new functional chemical compounds can bring enormous scientific and technological advances. For this reason, researchers in cheminformatics have developed a plethora of A.I. methodologies for the challenging inverse molecular design task \cite{sanchez2018inverse,coley2020}. They include deep learning techniques such as variational autoencoders (VAE) \cite{gomez2018automatic,jin2018junction,ma2018constrained}, generative adversarial networks (GAN) \cite{guimaraes2017objective,de2018molgan}, reinforcement learning (RL) \cite{zhou2019optimization,you2018graph}, and evolutionary techniques such as genetic algorithms (GA) \cite{jensen2019graph,nigam2019augmenting,henault2020chemical,reeves2020assessing}. 

These methods belong to a category with one particular attribute: the model \textit{indirectly} optimizes molecules for a target property. For example, VAEs and GANs learn to mimic a distribution of molecules from a training set, constructing a latent space that is then scanned to find molecules that optimize an objective function. In the case of RL, the agent learns from rewards in the environment in order to build a policy for generating molecules, which is subsequently used to maximize an objective function. Finally, in GAs, the population is optimized iteratively by applying mutations and selections. In all of these cases, the optimization process does not directly maximize the objective function in a gradient-based way.

Here, we present preliminary results for \Pasitheaa\describe, a new generative model for molecules inspired by inceptionism techniques \cite{mordvintsev2015inceptionism} in computer vision. \Pasithea is a gradient-based method that optimizes a discrete molecular structure for a target property. We train a neural network to predict chemical properties using a molecular string representation. We then invert the training of the network to generate new variants of molecules. This approach has two significant novelties: 
\begin{itemize}
	\item Molecules are \emph{directly} optimized to a given objective function, sidestepping the learning of distributions and policies, or the application of mutations to a population.
	\item We can analyse what the regression network has learned about the chemical property by probing its inverse training with test molecules. This may allow us to explain the neural network's understanding of chemistry.
\end{itemize}
Furthermore, in contrast to most exploratory methods such as RLs or GAs, \Pasithea does not require expensive function evaluations for quantum chemistry calculations. Provided that we use a pre-calculated dataset, this is an important advantage over explorative approaches such as GA or RL, since costly chemical properties can be directly optimized.

This method is made possible by the application of \selfies, a 100\% robust molecular string representation \cite{krenn2020self}. In contrast to SMILES, for which a large fraction of generable strings do not map to valid molecular graphs, \selfies is a surjective map between molecular strings and molecular graphs. That is, for every \selfies string, there exists a valid molecular graph, and every molecular graph can be represented by \selfies.

We train \Pasithea on the QM9 dataset \cite{ramakrishnan2014quantum} to predict logP values. We then initialize the inverse training with a molecule and optimize it for a logP value. We confirm a shift in the logP distribution of generated molecules. We can observe how the model changes a molecule quasi-continuously over several steps to a final, optimized chemical structure. Finally, we indicate how this technique can be used to probe concepts learned by \Pasithea. For completeness, we want to mention there are very recent technologies that can efficiently explore the chemical space that do not require training, datasets or domain-knowledge at all \cite{nigam_stoned}. Those techniques are complementary to what we demonstrate here, and it might be interesting to see how they can be combined with \Pasithea.

\section{Methodology}
Inceptionism \cite{simonyan2014very,mordvintsev2015inceptionism} has drawn considerable attention as an artistic method for rendering images. By using a neural network trained to classify an image (i.e., dog, car or house), the network can perform deep 'dreaming' on an image in order to mutate it gradually to fit a different class while retaining features of the original image. For example, it may enhance animal features in the image of a chemistry lab while the general structure of a lab is still visible (Figure \ref{fig:glassware}). The rendered images have dream-like properties that make them a popular artistic style in the media \cite{simonyan2014very}.

\begin{figure}[t]
\centering
\includegraphics[width=\textwidth]{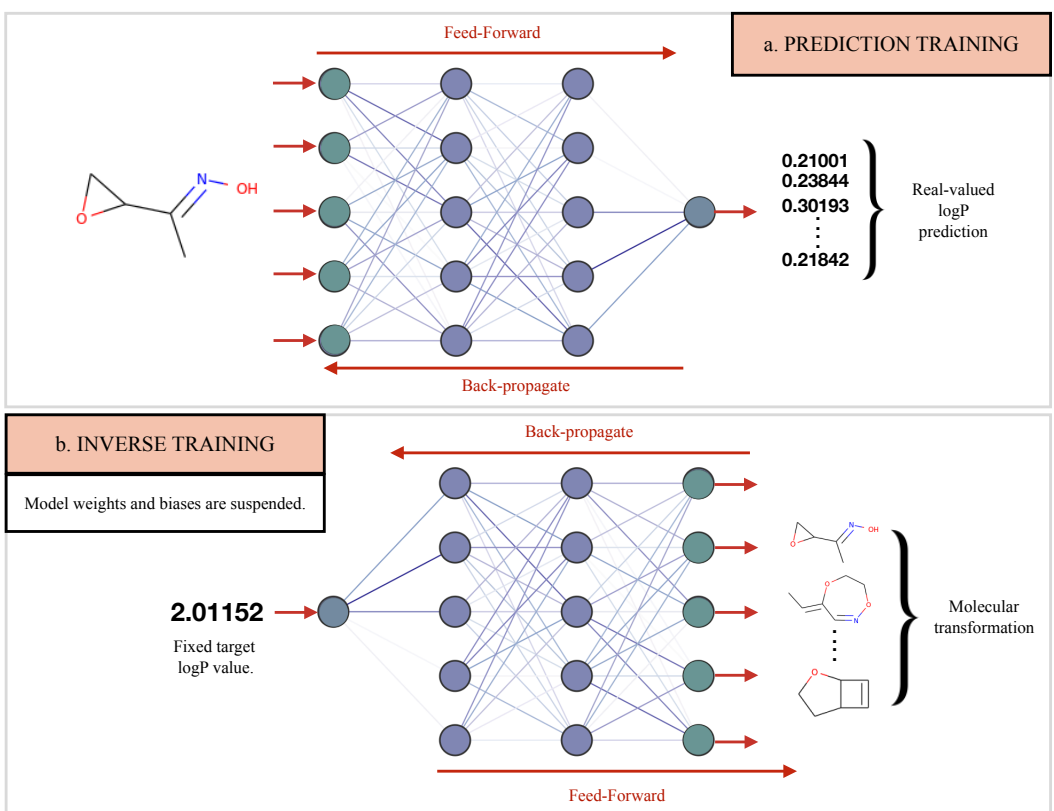}
\caption{Two-step training for \Pasithea. (a) In prediction training, the neural network learns to predict logP values from input molecules through the standard feedforward and backpropagation process in which the network weights are updated continually. (b) The same feedforward and backpropagation process occurs. The gradients are not computed with respect to the weights, but rather the input molecule.}
\label{fig:Concept}
\end{figure}

We generalize this methodology to the inverse-design task of functional molecules.  \Pasithea uses a fully-connected neural network consisting of four layers, each with 500 nodes, and takes as input the one-hot encoding of the SELFIES representation of each molecular graph. 

Prior to deep dreaming, the network learns to predict a specific real-valued property for each molecule in a given dataset (i.e., logarithm of partition coefficient, or logP) from the molecular graph. The training involves the standard feedforward and backpropagation process. For a set of fixed inputs and outputs, the network iteratively improves its predictions by updating the weights through mini-batch gradient descent (Figure \ref{fig:Concept}a). 
\begin{figure}[t]
\centering
\includegraphics[width=\textwidth]{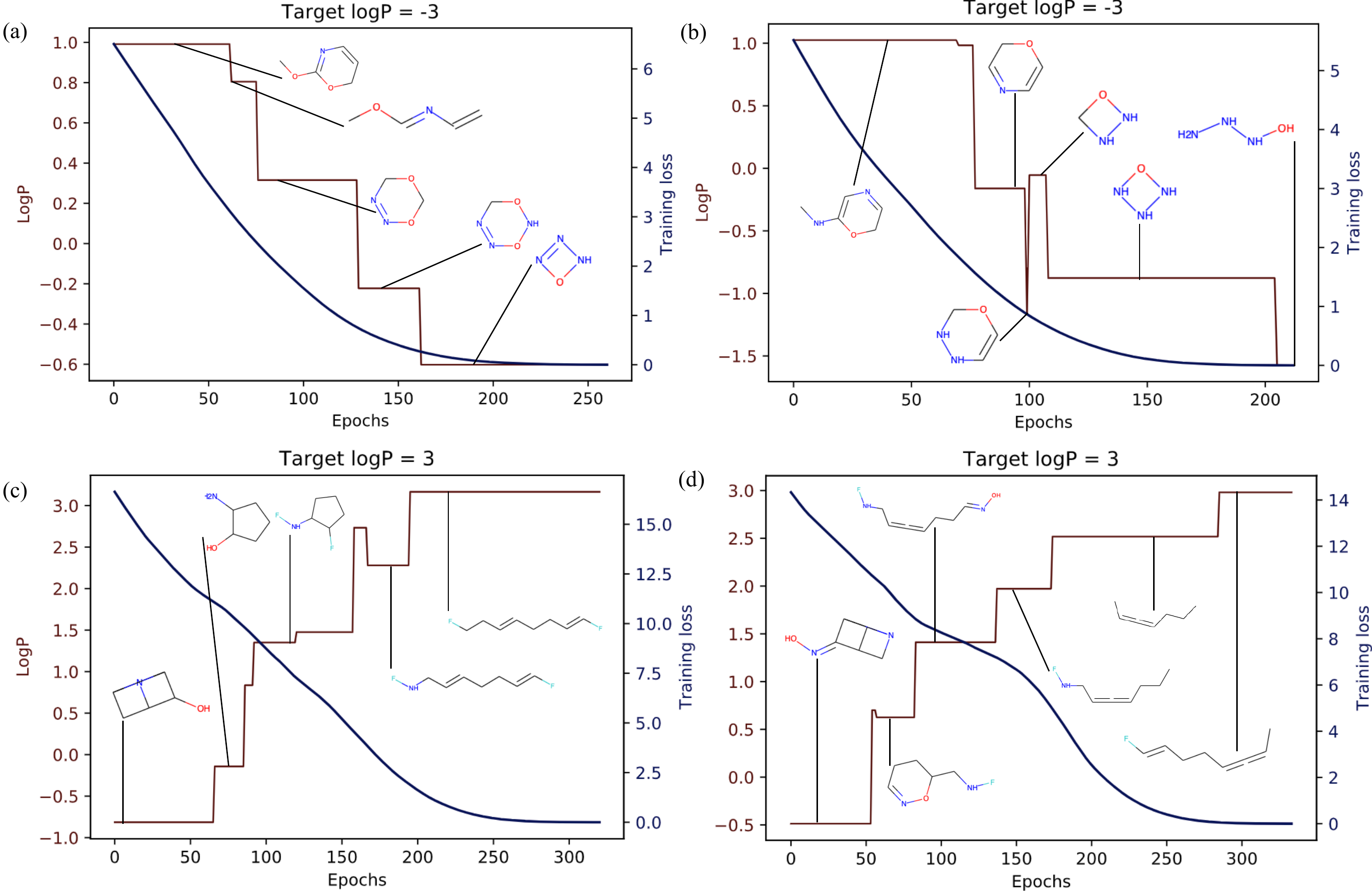}
\caption{Stepwise molecule transformations, optimized for a higher target logP, (a), and lower target logP, (b). In each plot, the blue curve represents the training loss over the number of epochs and the red curve represents logP adjustments corresponding to the training loss.}
\label{fig:evolution}
\end{figure}
In deep dreaming, an input molecule with a property value predicted by the network is incrementally modified to a similar molecule with the desired value. The weights and biases of each layer of the network are now fixed and the neural network is no longer adjusting its logP prediction for each molecule. Through backpropagation, we minimize the error between the predicted properties of each input molecule and the desired target property (Figure \ref{fig:Concept}b). The computed error is then used to compute the gradient with respect to the one-hot encoding of the input. This effectively transforms the input gradually to a molecule that matches the target property. Each increment of the one-hot encoding corresponds to a potential transformation of the input molecule. Once the loss function has been minimized, the gradient evaluates approximately to zero, which terminates the training. In this process, the same standard feedforward and backpropagation algorithm is used, but the input molecule is adjusted while the weights and biases remain constant. 
\section{Results}

It is an ongoing study to find the best numerical conversion from the \selfies string into an appropriate input for deep dreaming. When a one-hot encoding is taken as input, the encoding transforms from a vector of binary variables to a vector of real numbers in the first and subsequent iterations. As a result of mixed vector representations in training, the model has difficulty in converging. Therefore, taking one of several possible approaches, we introduce noise in the one-hot encoding as input to deep dreaming. Every zero in the one-hot encoding is altered to a random number between zero and a specific upper-bound, which is typically set to a value between 0.5 and 0.9. Using this method, we observe an incremental optimization for each given molecular input, as required.

Another important contribution to the model is the application of \selfies. This method requires a continuous space in which all points are valid, a criterion met by the recently developed \selfies, which is proven to be 100 \% valid \cite{krenn2020self}. The traditional SMILES representation can be problematic when the deep dreaming model transitions over an invalid structure from one molecule to the next. For example, in the transition from a string containing a ring,"CCCC1CCCC1CC", to a string without a ring, "CCCCCCCCCC", the model is likely to produce strings resembling "CCCC1CCCCCC", which does not correspond to a valid molecular graph. In this case, the transformation may reveal the network's understanding of string syntax in relation to logP, but not the molecular structure in relation to logP, since the string does not correspond to a valid molecule. In contrast, the \selfies representation enforces a constraint on the syntax to prevent the model from producing such invalid structures, which produces a complete optimization sequence that directly maps to valid molecules. Our findings here highlight only one of the many potential applications of \selfies.

\begin{figure}[t]
\centering
\includegraphics[width=\textwidth]{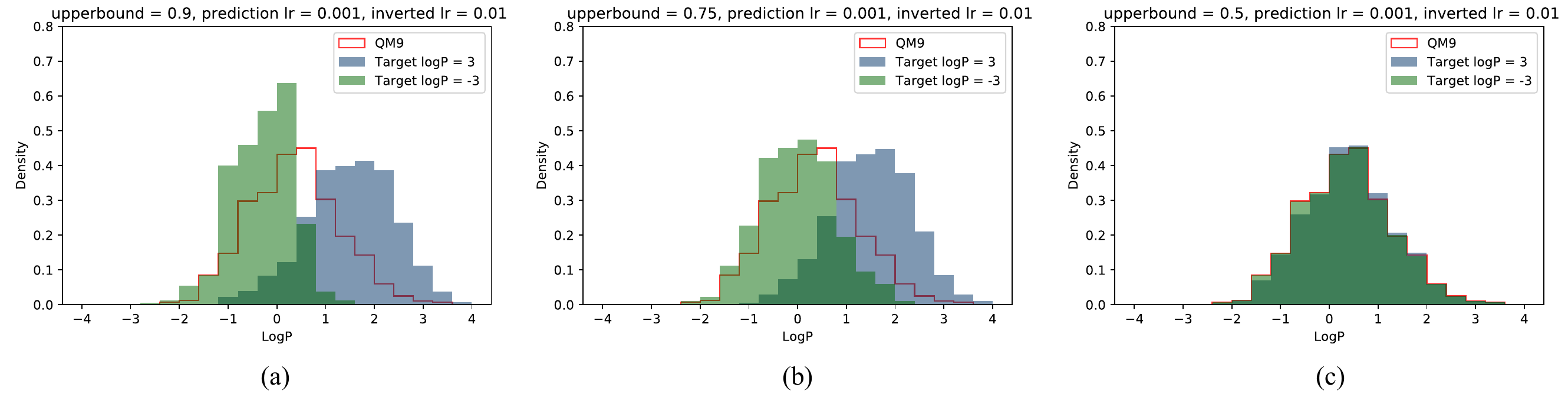}
\caption{Shifts in distribution. a) At an upper-bound of 0.9, the molecules are optimized toward logP targets at both extreme ends. c) At a substantially lower upper-bound, the model is no longer capable of optimizing the molecules.}
\label{fig:distribution}
\end{figure}
Our experiments clearly indicate that deep dreaming achieves both a direct, gradient-based design of novel functional molecules and the explainability of neural networks for molecules. A simple four-layer neural network, with no added components, suffices for our results and we did not require an exhaustive search for the ideal training hyperparameters. In this analysis, \Pasithea is trained to predict the logarithm of partition coefficient (logP), obtained from the RDKit library \cite{landrum2006rdkit}, on a set of the smallest 10,000 molecules in the QM9 dataset. The logP, which measures the lipophilicity of a molecule, is an important property of drug molecules and an indicator of drug-likeness \cite{lipinski2004lead}. We demonstrate how \Pasithea transforms molecules in a stepwise, quasi-continuous fashion and shifts the distribution of logP in the molecular dataset toward set targets. These logP targets are set high in order to observe a rightward shift in logP distribution and similarly set low to observe a shift in the opposite direction. With logP targets much further from the central tendency in the distribution, surpassing the highest and lowest values in the dataset, we observe a more pronounced shift in distribution during training. We then analyse what \Pasithea has learned regarding the relationship between logP and molecular structure.

\subsection{Evolution of individual molecules}
Of particular interest is the gradual progression of each molecule through inverse training. Over hundreds of training epochs, the gradient with respect to input \selfies produces minor adjustments in the molecule that increments to a pronounced transmutation (Figure \ref{fig:evolution}). The behaviour of these adjustments are stepwise due to the discrete, textual nature of the molecules represented by strings, but continuous in terms of real-valued one-hot encodings.

\begin{figure}[h]
\centering
\includegraphics[width=\textwidth]{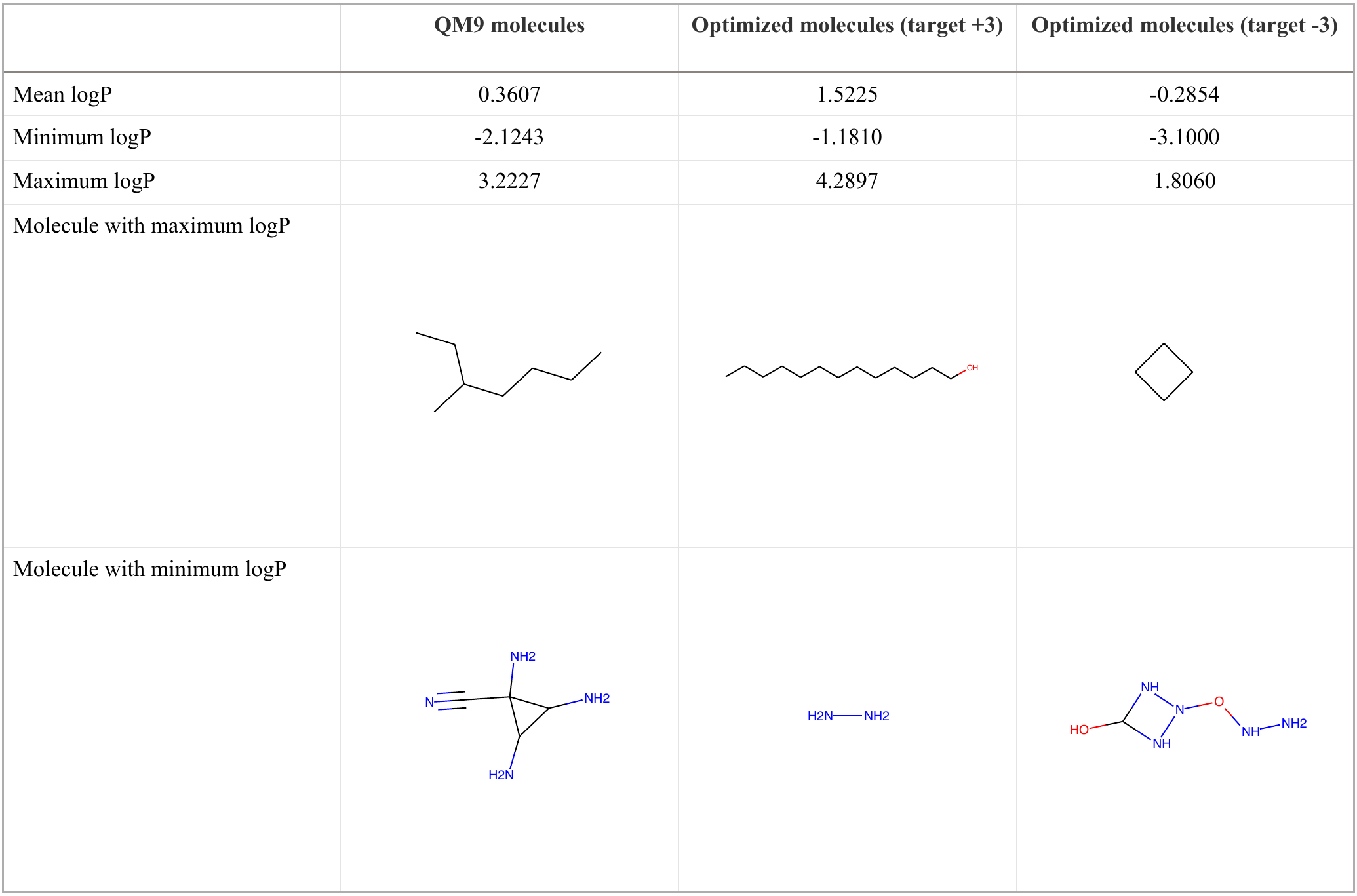}
\caption{The mean, minimum and maximum corresponding to the distribution in Figure \ref{fig:distribution}a.}
\label{fig:stats}
\end{figure}

\subsection{Shift in distribution}
In order to observe a large-scale pattern over the entire dataset, we disregard the intermediate molecules and restrict our analysis to the initial and fully-optimized molecules. We take a sample of the smallest 10,000 molecules from the QM9 dataset and apply deep dreaming to each molecule. From these results, there is a clear shift in the distribution of logP values in the set of molecules as they transmute toward a given target value (Figure \ref{fig:distribution}). Although the training learning rates have little effect on the quality of training, the addition of more noise to one-hot encoded inputs (higher upper-bound values in Figure \ref{fig:distribution}) has a large influence on the shifts in distribution curves. We furthermore observe from these distribution shifts in Figure \ref{fig:distribution} that there are some molecules generated with logP values exceeding the lowest and highest values in the original dataset. For instance, notice that in Figure \ref{fig:distribution}a, the left tail of the left-shifted (green) distribution extends beyond the left tail of the original (red) distribution and the right tail of the right-shifted (blue) distribution extends beyond the right tail of the original distribution. A quantitative account of the distribution in Figure \ref{fig:distribution}a is summarized in Figure \ref{fig:stats}. Notice that the maximum logP in the right-shifted distribution exceeds that in the original dataset, and the minimum in the original dataset exceeds that in the left-shifted distribution. Demonstrably, \Pasithea is generating novel molecules with properties outside the limits of the original training set of molecules, which attests to the large potential of this method.

\subsection{Probing the neural network's intuitions}
\subsubsection{Interpretable ML in physics}
Our approach to de-novo molecular generation does not require domain knowledge, nor is the design of \Pasithea influenced by domain knowledge. However, this knowledge is useful when applied to individual molecular evolutions in deep dreaming \cite{8466590}. In particular, we take interest in the recent progress in the machine-assisted discovery of concepts in the natural sciences \cite{seif2020machine,PhysRevLett.124.010508,roscher2020explainable,friederich2020scientific}. These lines of research use machine learning techniques to draw conclusions about the underlying processes of a particular physical system, which are often mathematical models with tunable parameters that are responsive to input observations. This approach differs from research in machine-assisted de-novo molecular generation \cite{gomez2018automatic}, where the focus lies in producing optimization methods that can navigate a massive chemical search space. Our approach may close the gap between these lines of research. By inverting the training, we achieve both molecular generation and insights into how the network produces each molecular transformation, such as the 'strategy' employed to optimize molecules by appending nitrogen atoms. Although \Pasithea does not model the behaviour of molecules in the physical sense, it does model the transition rules required for molecular optimization; there is potential in rigorously quantifying these transition rules. Specifically, the viability of inceptionism in recovering the thermodynamic principles of physics \cite{seif2020machine} attests to the potential for chemistry.
\subsubsection{Interpretable ML in chemistry}
\begin{figure}[t]

\centering
\includegraphics[width=\textwidth]{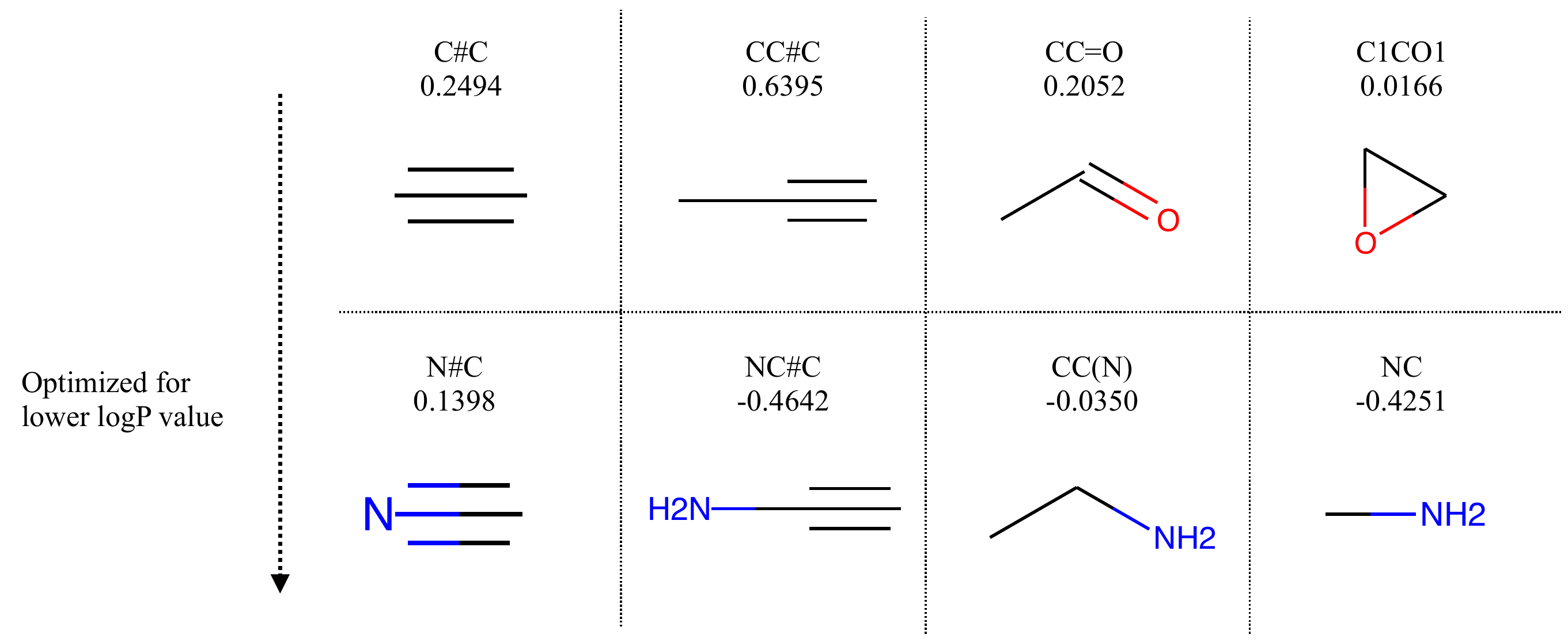}
\caption{Graphical and SMILES representation of nitrogen appendages, with the corresponding logP. Shown in the top row are initial test molecules from the QM9. Shown in the bottom row are the optimized counterparts of the molecules directly above. The neural network recognizes the nitrogen atom as an important indicator of a low logP value.}
\label{fig:appendage}
\end{figure}

Inspired by explainable representations in image recognition \cite{mahendran2015understanding} and rediscovery of concepts in physics \cite{seif2020machine}, we can \textit{understand the internal molecular representation by inverting it}. For that, we probe the neural network with specific test molecules and observe patterns in how it changes them.
For example, the composition of atoms after inverse training follows a predictable pattern, such as the appendage of a few non-carbon atoms, fluorine and nitrogen. Take, for example, the transmutations of the simplest molecules in the QM9 dataset (Figure \ref{fig:appendage}), which suggest that \Pasithea interprets these non-carbon atoms as correlated with lower logP values. A similar trend persists for more complex molecules, in which more than one atom may be replaced with nitrogen (Figure \ref{fig:evolution}a), though this persists to a lower extent for fluorine. 

The intermediate states during continuous transformation can be used as additional insights into the network's understanding of chemical property. In particular, by observing a single test molecule, there are instances where an additional iteration in inverse-training transforms the molecule with a repeated 'strategy' that has been used in previous iterations. The neural network appears to persist with a single strategy until the training terminates. We demonstrate this behaviour in Figure \ref{fig:evolution}b, which shows a gradual process of reducing length, and in Figure \ref{fig:evolution}a, which shows an initial molecule containing a single nitrogen atom, an intermediate molecule containing two nitrogen atoms, and a final molecule containing three nitrogen atoms. These cases validate that the network is charting deliberate, non-arbitrary paths toward the target logP; it has a non-trivial understanding of features corresponding to higher and lower logP values.

\section{Comparison to VAEs}
A simple 4-layer network highlights one key difference between \Pasithea and other optimization methods: we perform reverse-differentiation directly on the molecular representation, which is a one-hot encoding of \selfies. Let us compare this approach with the related concept of variational autoencoder \cite{gomez2018automatic}. In VAEs, a latent space is learned by encoding and decoding molecules. After the reconstruction, another neural network can then optimize in the newly created latent space. In this case, since the prediction network is applied to the latent space, the basis for gradient computation lies in the latent space, not in the molecular representation itself.

The direct reversibility on the basis of model weights is important in the context of machine learning interpretability. Our goal is to understand directly what a neural network learns about a specific molecular property. We believe that probing the regression neural network with test molecules, without a detour over some specific latent spaces, is the most direct way to understand what the model has learned.  

\section{Outlook} 
We propose a direct, gradient-based property-optimization method that offers insights into the network's understanding of structure-property relationships. In the immediate future, we will verify our results on larger datasets and more complex molecules, such as PubChem. Furthermore, we plan to test \Pasithea on molecular properties that require expensive quantum chemistry calculations. We see much potential in discovering other 'strategies' the network may use in order to optimize molecules with different properties.

There is also work to be done to add transparency \cite{8466590} to our approach. There are many possible directions, including exploring other surjective string representations that may be more suitable to the task of deep dreaming, and comparing other reverse-differentiable machine learning architectures that may be capable of a similar 'dreaming' process. Ultimately, our work can be used to find the underlying rules the neural network discovers in order to optimize a property, conjointly offering insights into how the network makes its predictions for interpretability and suggesting ways in which a human can use these rules in order to generate new and useful chemical compounds for explainability \cite{8466590}.

\section{Acknowledgements}
A. A.-G. acknowledges generous support from the Canada 150 Research Chair Program, Tata Steel, Anders G Froseth, and the Office of Naval Research. M.K. acknowledges support from the Austrian Science Fund (FWF) through the Erwin Schr\"odinger fellowship No. J4309.

\bibliographystyle{unsrt}
\bibliography{refs}

\begin{thebibliography}{}

\bibitem{krenn2020self}
M. Krenn, F. Hase, A. Nigam, P. Friederich and A. Aspuru-Guzik,
  Self-Referencing Embedded Strings (SELFIES): A 100\% robust molecular string
  representation. \textit{Machine Learning: Science and Technology} \textbf{1},
  045024 (2020).

\bibitem{mordvintsev2015inceptionism}
A. Mordvintsev, C. Olah and M. Tyka, Inceptionism: Going deeper into neural
  networks.
  \textit{\url{https://ai.googleblog.com/2015/06/inceptionism-going-deeper-into-neural.html}}
  (2015).

\bibitem{simonyan2014very}
K. Simonyan and A. Zisserman, Very deep convolutional networks for large-scale
  image recognition. \textit{arXiv:1409.1556} (2014).

\bibitem{linden2019github}
E. Linder-Noren, Pytorch - Deep Dream.
  \textit{\url{https://github.com/eriklindernoren/PyTorch-Deep-Dream}} (2019).

\bibitem{sanchez2018inverse}
B. Sanchez-Lengeling and A. Aspuru-Guzik, Inverse molecular design using
  machine learning: Generative models for matter engineering. \textit{Science}
  \textbf{361}, 360--365 (2018).

\bibitem{coley2020}
C.W. Coley, Defining and Exploring Chemical Spaces. \textit{Trends in
  Chemistry} (2020).

\bibitem{gomez2018automatic}
R. G{\'o}mez-Bombarelli, J.N. Wei, D. Duvenaud, J.M. Hern{\'a}ndez-Lobato, B.
  S{\'a}nchez-Lengeling, D. Sheberla, J. Aguilera-Iparraguirre, T.D. Hirzel,
  R.P. Adams and A. Aspuru-Guzik, Automatic chemical design using a data-driven
  continuous representation of molecules. \textit{ACS central science}
  \textbf{4}, 268--276 (2018).

\bibitem{jin2018junction}
W. Jin, R. Barzilay and T. Jaakkola, Junction tree variational autoencoder for
  molecular graph generation. \textit{arXiv:1802.04364} (2018).

\bibitem{ma2018constrained}
T. Ma, J. Chen and C. Xiao, Constrained generation of semantically valid graphs
  via regularizing variational autoencoders. \textit{Advances in Neural
  Information Processing Systems} 7113--7124 (2018).

\bibitem{guimaraes2017objective}
G.L. Guimaraes, B. Sanchez-Lengeling, C. Outeiral, P.L.C. Farias and A.
  Aspuru-Guzik, Objective-reinforced generative adversarial networks (ORGAN)
  for sequence generation models. \textit{arXiv:1705.10843} (2017).

\bibitem{de2018molgan}
N. De~Cao and T. Kipf, MolGAN: An implicit generative model for small molecular
  graphs. \textit{arXiv:1805.11973} (2018).

\bibitem{zhou2019optimization}
Z. Zhou, S. Kearnes, L. Li, R.N. Zare and P. Riley, Optimization of molecules
  via deep reinforcement learning. \textit{Scientific reports} \textbf{9},
  1--10 (2019).

\bibitem{you2018graph}
J. You, B. Liu, Z. Ying, V. Pande and J. Leskovec, Graph convolutional policy
  network for goal-directed molecular graph generation. \textit{Advances in
  neural information processing systems} 6410--6421 (2018).

\bibitem{jensen2019graph}
J.H. Jensen, A graph-based genetic algorithm and generative model/Monte Carlo
  tree search for the exploration of chemical space. \textit{Chemical science}
  \textbf{10}, 3567--3572 (2019).

\bibitem{nigam2019augmenting}
A. Nigam, P. Friederich, M. Krenn and A. Aspuru-Guzik, Augmenting genetic
  algorithms with deep neural networks for exploring the chemical space.
  \textit{arXiv:1909.11655} (2019).

\bibitem{henault2020chemical}
E.S. Henault, M.H. Rasmussen and J.H. Jensen, Chemical space exploration: how
  genetic algorithms find the needle in the haystack. \textit{PeerJ Physical
  Chemistry} \textbf{2}, e11 (2020).

\bibitem{reeves2020assessing}
S. Reeves, B. DiFrancesco, V. Shahani, S. MacKinnon, A. Windemuth and A.E.
  Brereton, Assessing Methods and Obstacles in Chemical Space Exploration.
  \textit{Applied AI Letters} ail2.17 (2020).

\bibitem{ramakrishnan2014quantum}
R. Ramakrishnan, P.O. Dral, M. Rupp and O.A. Von~Lilienfeld, Quantum chemistry
  structures and properties of 134 kilo molecules. \textit{Scientific data}
  \textbf{1}, 1--7 (2014).

\bibitem{nigam_stoned}
A. Nigam, R. Pollice, M. Krenn, G. Passos~Gomes and A. Aspuru-Guzik, Beyond
  Generative Models: Superfast Traversal, Optimization, Novelty, Exploration
  and Discovery (STONED) Algorithm for Molecules using SELFIES.
  \textit{ChemRxiv 13383266} (2020).

\bibitem{landrum2006rdkit}
G. Landrum and  others, RDKit: Open-source cheminformatics. (2006).

\bibitem{lipinski2004lead}
C.A. Lipinski, Lead-and drug-like compounds: the rule-of-five revolution.
  \textit{Drug Discovery Today: Technologies} \textbf{1}, 337--341 (2004).

\bibitem{8466590}
A. {Adadi} and M. {Berrada}, Peeking Inside the Black-Box: A Survey on
  Explainable Artificial Intelligence (XAI). \textit{IEEE Access} \textbf{6},
  52138-52160 (2018).

\bibitem{seif2020machine}
A. Seif, M. Hafezi and C. Jarzynski, Machine learning the thermodynamic arrow
  of time. \textit{Nature Physics} 1--9 (2020).

\bibitem{PhysRevLett.124.010508}
R. Iten, T. Metger, H. Wilming, L. Rio and R. Renner, Discovering Physical
  Concepts with Neural Networks. \textit{Phys. Rev. Lett.} \textbf{124}, 010508
  (2020).

\bibitem{roscher2020explainable}
R. Roscher, B. Bohn, M.F. Duarte and J. Garcke, Explainable machine learning
  for scientific insights and discoveries. \textit{IEEE Access} \textbf{8},
  42200--42216 (2020).

\bibitem{friederich2020scientific}
P. Friederich, M. Krenn, I. Tamblyn and A. Aspuru-Guzik, Scientific intuition
  inspired by machine learning generated hypotheses. \textit{arXiv:2010.14236}
  (2020).

\bibitem{mahendran2015understanding}
A. Mahendran and A. Vedaldi, Understanding deep image representations by
  inverting them. \textit{Proceedings of the IEEE conference on computer vision
  and pattern recognition} 5188--5196 (2015).

\end{thebibliography}

\end{document}